\documentclass[preprint,5p]{elsarticle}

\usepackage{amssymb}
\usepackage{amsmath}

\usepackage{multirow}
\usepackage{tabularx}
\usepackage[table]{xcolor}
\definecolor{table_1blue}{HTML}{C0C0C0}   
\definecolor{table_2blue}{HTML}{E0E0E0}  
\journal{Medical Imaging Analysis}

\begin{document}

\begin{frontmatter}



\title{Learning Robust and Task-Invariant Functional Representation from fMRI through Siamese Self-Supervised Learning}


\author[bme]{Jiyao Wang} %% Author name
\author[bme]{Peiyu Duan} %% Author name
\author[bme,rad]{Nicha C. Dvornek} %% Author name
\author[bme,rad,ee]{Lawrence H. Staib} %% Author name
\author[csc]{Denis Sukhodolsky} %% Author name
\author[csc]{Pamela Ventola} %% Author name
\author[bme,rad,ee]{James S. Duncan} %% Author name

\affiliation[bme]{organization={Department of Biomedical Engineering},
            addressline={Yale University}, 
            city={New Haven},
            state={CT},
            country={USA}}

\affiliation[rad]{organization={Radiology \& Biomedical Imaging},
            addressline={Yale School of Medicine}, 
            city={New Haven},
            state={CT},
            country={USA}}

\affiliation[ee]{organization={Electrical Engineering},
            addressline={Yale University}, 
            city={New Haven},
            state={CT},
            country={USA}}

\affiliation[csc]{organization={Child Study Center},
            addressline={Yale School of Medicine}, 
            city={New Haven},
            state={CT},
            country={USA}}

%% Abstract
\begin{abstract}
Functional magnetic resonance imaging (fMRI) is a powerful tool for investigating human brain function. However, the high cost of data acquisition and the inherent subjectivity of psychiatric rating scales often lead to datasets with small sample sizes and variable label quality, especially when targeting a specific neurological condition. Combined with the inherently high dimensionality of fMRI data, these limitations substantially increase the risk of model overfitting. Recent years have seen growing interest in developing fMRI foundation models by combining multiple datasets; however, the computational resources needed for pretraining and fine-tuning are often prohibitive. We show that a lightweight self-supervised framework yields representations that generalize across diverse downstream tasks, outperforming fully supervised baselines and approaching the performance of large-scale models. We introduce BrainSimSiam, a data-efficient self-supervised representation learning framework that leverages positive-only data pairs to learn robust and generalizable features. We demonstrate that the learned representations achieve strong performance across multiple downstream classification and regression tasks, highlighting the potential of BrainSimSiam for data-limited neuroimaging applications.
\end{abstract}


% \begin{highlights}
% \end{highlights}

%% Keywords
\end{frontmatter}

\section{Introduction}
\label{sec:introduction}
Through the integration of deep learning and functional magnetic resonance imaging (fMRI), significant advances have been made in understanding neurological and psychiatric disorders such as Alzheimer’s disease~\cite{alzheimer}, attention deficit hyperactivity disorder (ADHD) \cite{adhd}, and autism spectrum disorder (ASD) \cite{braingnn}. Although these disorders are inherently heterogeneous~\cite{alz-sub, adhd-sub, asd-sub}, most supervised learning frameworks often rely on binary classification labels. Collapsing diverse disease subtypes and healthy controls (HC) into binary categories can obscure meaningful subtype differences, potentially leading to model overfitting and reduced interpretability.
%Unsupervised learning approaches are potentially more appropriate for modeling the distributions for not holding the binary assumption.

% noisy label for machine learning
In addition, labels for neurological disorders are inherently noisy due to variability in disease severity and human factors in assessment. Although machine learning has driven substantial gains in medical imaging, models usually remain vulnerable to noise in labels~\cite{noisy-survey}. In supervised learning settings, even with strong augmentation and regularization techniques~\cite{dvornek2018,braingnn}, overfitting can produce distorted decision boundaries, limiting the generalization capability of models. Label-independent approaches, especially unsupervised and self-supervised learning, offer a promising path to more robust representation learning and downstream performance.

% brain + deep learning: 1) supervised 2) unsupervised
A wide range of machine learning architectures have been proposed for fMRI analysis, covering supervised~\cite{dvornek2018,braingnn,stnagnn} and self-supervised~\cite{contrastive-augmentation,contrastive-population,simese-rep,contrastive-stgcn,zhou2024self} paradigms. Despite this architectural diversity, most methods consider fMRI as a homogeneous sequence during acquisition for both resting-state and task-based data. In recent years, growing evidence indicates that task-based fMRI provides richer, behaviorally grounded information than resting-state fMRI data~\cite{vince-task,behav-task}. Moreover, the functional variation induced by diverse fMRI task paradigms yields informative contrasts that can be leveraged to learn robust, task-invariant functional representations. For data-limited applications, truncating task-based fMRI sequences into task-homogeneous subsequences also increases the effective sample size.

% contributions
In this work, we introduce Brain Simple Siamese representation learning~(BrainSimSiam), a lightweight and data-efficient self-supervised framework for representation learning of fMRI. In a two-step training scheme, we show that the pretrained BrainSimSiam encoder produces more robust discriminative models in both multilayer perceptron (MLP) probing and end-to-end finetuning settings for downstream applications. Experiments are performed on classification and regression tasks including predictions of both biological and behavioral phenotypes. Our contributions are:
\begin{itemize}
    \item We develop a self-supervised representation learning framework for task-based fMRI that outperforms supervised and self-supervised baselines on both datasets examined.
    \item We unify voxel-wise fMRI and graph-based functional views via a joint region of interest (ROI) masking scheme applied during training and post hoc interpretation. The same mechanism readily supports multimodal fusion by incorporating registered structural T1-weighted or diffusion MRI with the functional brain graph.
    \item We leverage contrasts across diverse task stimuli as natural augmentations to learn task-invariant representations, disentangling evoked responses from spontaneous brain dynamics.
\end{itemize}

\section{Related Work}
% GNN
\subsection{Graph Neural Network~(GNN)}
GNNs are a family of neural networks for graph-structured data~\cite{gcn,gat,sage}. In ROI-based brain graphs, nodes correspond to regions of interest and edges encode functional or structural connectivity~\cite{braingnn,stnagnn,zhao-stgnn}. GNNs learn by iteratively aggregating information from each node’s neighbors. Let $h_i^{k}$ denote the feature of node $i$ at layer $k$ and $\mathcal{N}(i)$ denote its neighbor set. A generic message-passing update of a GNN can be described as 
\begin{equation}
h_i^{k+1} = \phi \left(h_i^{k}, m_{ij, j\in \mathcal{N}(i)}| m_{ij}=\psi\big(h_i^{k},h_j^{k},e_{ij}\big) \right)
\end{equation}
where $\psi$ computes messages $m$, $e_{ij}$ are edge attributes, and $\phi$ produces the updated node representation. 

% SimSiam
\subsection{Simple Siamese~(SimSiam)}
Contrastive frameworks such as MoCo v2~\cite{mocov2} and SimCLR~\cite{simclr} learn invariances by pulling together augmented views of the same sample while pushing apart others. They have shown greater robustness to label noise and spurious correlations than supervised pretraining, but typically rely on large numbers of negatives via large batches or memory queues. SimSiam~\cite{simsiam} removes the need for negatives by using a stop-gradient Siamese architecture with a prediction head, achieving competitive performance in natural image applications with substantially simpler training. This property is also advantageous in neuroimaging settings characterized by modest batch sizes, where contrastive methods that depend on large negative sets are operationally fragile.

Although negative examples can be informative, defining generalizable negatives in fMRI is challenging. Treating different subjects as negatives presumes inter-subject dissimilarity and discards shared functional patterns. Constructing negatives from task labels entangles features with the label space, leading to oversimplified and task-bound representations, undermining cross-task generalization. 

We therefore adopt SimSiam~\cite{simsiam} as our framework for fMRI representation learning, a more suitable choice for learning generalizable representations across downstream tasks.

\section{Data and Preprocessing}
\subsection{Human Connectome Project (HCP) Dataset}
We use task-based fMRI scans from the WU–Minn Human Connectome Project (HCP) S1200 young-adult release~\cite{hcp} with seven task paradigms: emotion, gambling, language, motor, relational, social, and working memory. All images were spatially normalized to the MNI152 standard space and resampled to 2 mm isotropic voxels. After excluding subjects with missing task runs or behavioral measures, the final cohort comprises 1,005 participants (Table~\ref{tab_hcp}). Evaluation is conducted with 5-fold cross-validation with equal numbers per fold.

% table dim
\begin{table}[t]
\centering
\setlength{\tabcolsep}{8pt}
\begin{tabularx}{\columnwidth}{l c | c c }
\multicolumn{2}{l|}{} & Number & Percentage\\ 
\hline
\hline
\multicolumn{2}{l|}{\textbf{All Subjects}} & 1005 & 100\%\\ 
\hline
\textbf{\multirow{2}*{Biological Sex}} & Female  & 535 & 53\%\\ 
& Male  & 470 & 47\%\\ 
\hline
\textbf{\multirow{4}*{Age (years)}} & 22-25 & 216 & 21\%\\ 
& 26-30 & 438 & 44\%\\ 
& 31-35 & 341 & 34\%\\ 
& 36+  & 10 & 1\%\\ 
\hline

\end{tabularx}
\caption{Statistics of subjects in the HCP dataset}
\label{tab_hcp}
\end{table}

\subsection{Biopoint Dataset}
% table dim
\begin{table}[t]
\centering
\setlength{\tabcolsep}{14pt}
\begin{tabularx}{\columnwidth}{l c | c c }
\multicolumn{2}{l|}{} & ASD & HC\\ 
\hline
\hline
\multicolumn{2}{l|}{\textbf{All Subjects}} & 75 & 43 \\ 
\hline
\textbf{\multirow{2}*{Biological Sex}} & Female  & 12 & 20 \\ 
& Male  & 63 & 23 \\ 
\hline
\textbf{\multirow{2}*{Age (years)}} & Mean  & 9.50 & 10.92 \\ 
& Std  & 3.97 & 3.07 \\ 
\hline
\end{tabularx}
\caption{Statistics of ASD and HC subjects in the Biopoint dataset}
\label{tab_biopoint}
\end{table}

\begin{figure}
\centering
\includegraphics[width=0.8\columnwidth]{ 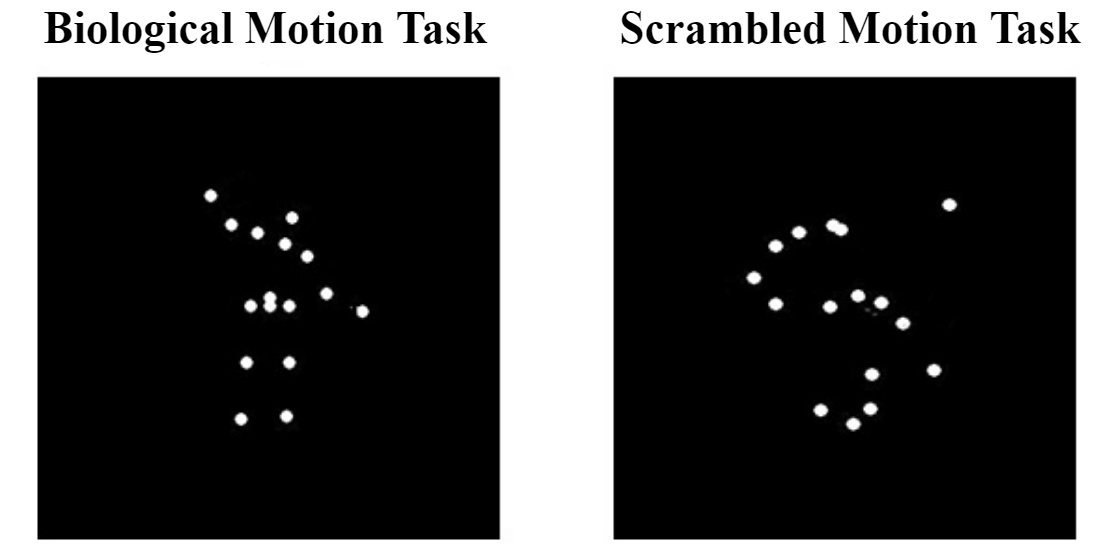}
\caption{Example of biological and scrambled motion tasks in the Biopoint dataset} \label{fig_biopoint}
\end{figure}

\begin{figure}
\centering
\includegraphics[width=\columnwidth]{ 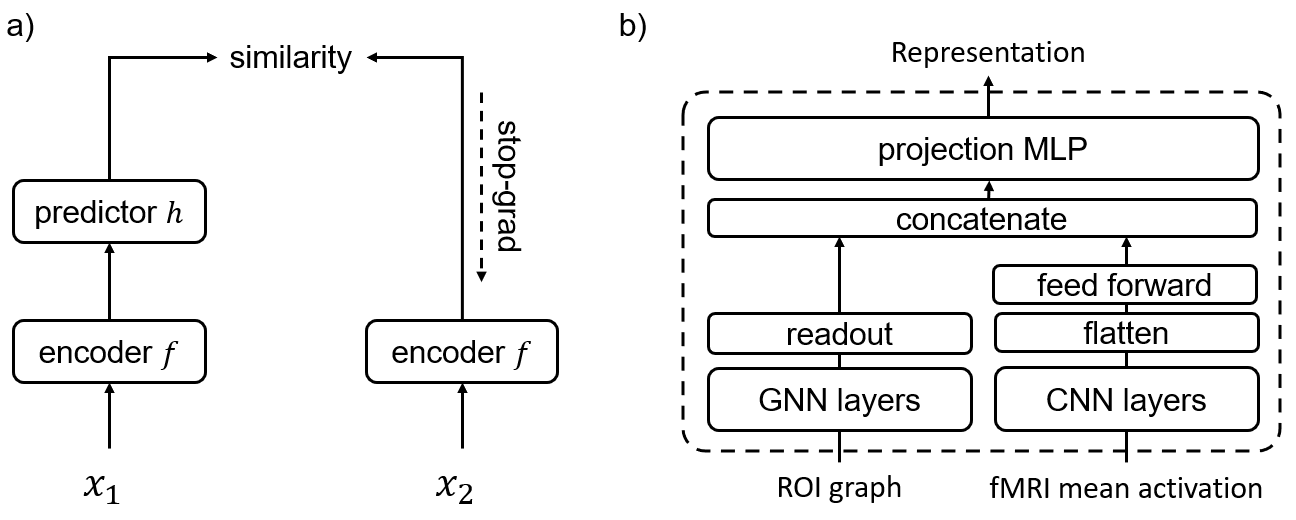}
\caption{Model architectures. (a) General SimSiam framework. (b) BrainSimSiam encoder $f$.} \label{fig_encoder}
\end{figure}

The Biopoint dataset~\cite{biopoint} comprises task-based fMRI acquired in an ASD study using 12 point-light display videos as task stimuli. Each 24-second video falls into one of two categories: biological motion depicting socially meaningful human actions and scrambled motion with random point trajectories ~(Fig.~\ref{fig_biopoint}). Videos from two tasks are presented in alternating blocks during acquisition with the intention of highlighting deficits in the perception of biological motion in autistic children. 

The preprocessing follows Yang et al.~\cite{biopoint-preprocess}, with images normalized to MNI152 standard space and resampled to 2 mm isotropic voxels. The dataset includes 75 children with ASD and 43 healthy controls (HC) matched in age and cognitive ability. For evaluation, we use a stratified 5-fold cross-validation to maintain similar ASD/HC proportions in each fold. For each subject, we truncate the Biopoint fMRI sequence into 12 blocks aligned with the 12 video stimuli. Each 24-second block is treated as a distinct task in our representation learning experiments. In Table~\ref{tab_biopoint}, we show the biological sex and age statistics of subjects in the Biopoint dataset.

\begin{figure*}
\centering
\includegraphics[width=\textwidth]{ 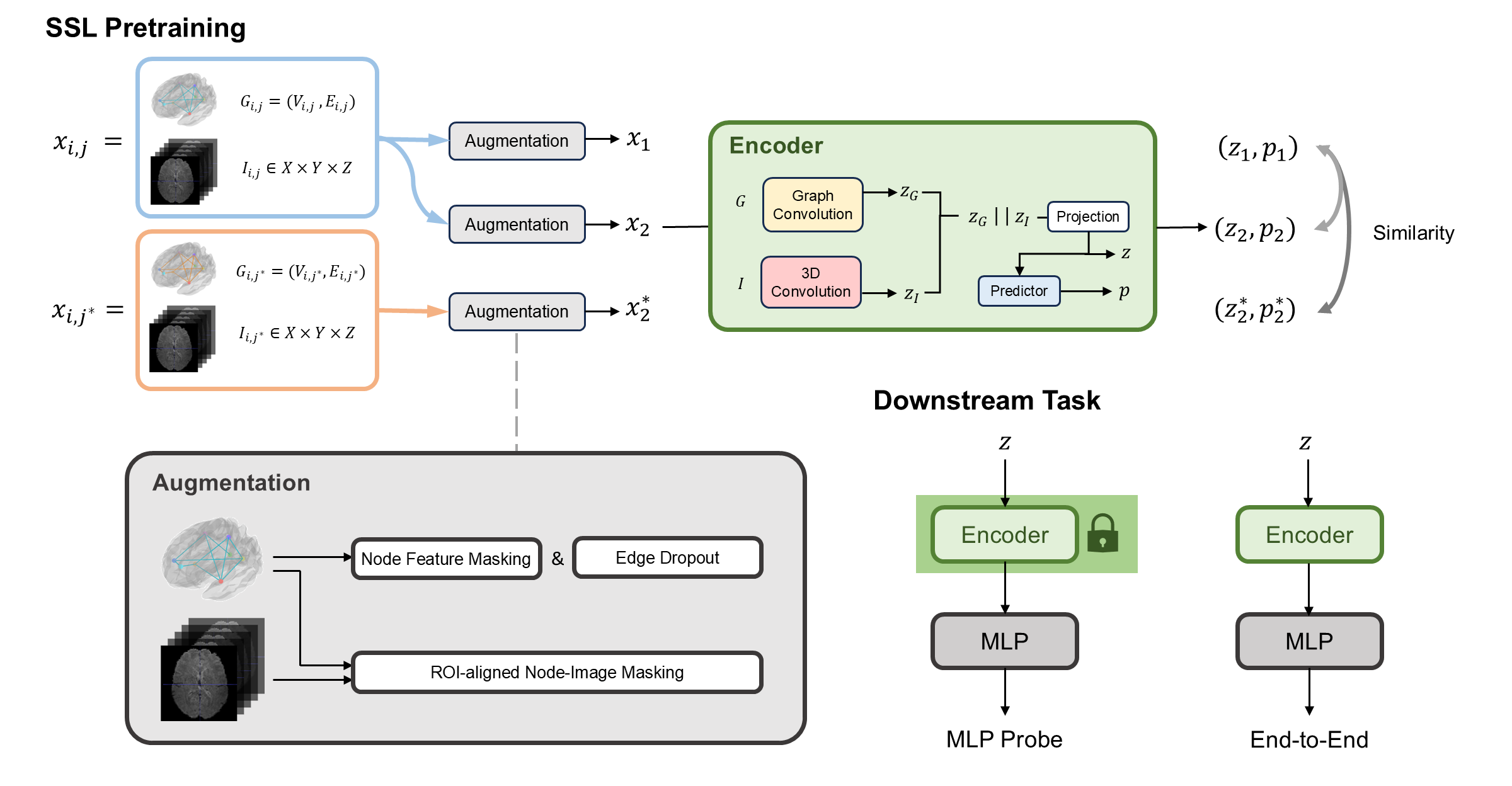}
\caption{Pipeline of the BrainSimSiam pretraining and downstream application} \label{fig_pipeline}
\end{figure*}

\subsection{Graph Construction and Image Preprocessing}

In the HCP dataset, each task instance corresponds to a full task-based fMRI run. The total number of data instances is $7035=1005\times7$. In the Biopoint dataset, due to the limited sample size, we define a task instance as each non-overlapping sliding-window subsequence of the fMRI time series acquired during a single video stimulus, leading to $1416=118\times12$ data instances. For each data instance, we derive two complementary views of the fMRI data, described below.

\subsubsection{Functional Brain Graph View} For the functional brain graph, each fMRI series is parcellated into ROIs using the Shen atlas (268 ROIs) for HCP and the Desikan–Killiany atlas (84 ROIs) for Biopoint~\cite{shen268,desikan}. Following the brain graph construction method presented in Li et al.~\cite{braingnn}, we compute both Pearson and partial correlations between ROI time series. Node features are defined as the corresponding rows of the Pearson correlation matrix, while the edges are retained as the strongest 5\% absolute value of the partial-correlation coefficients between ROIs. 

\subsubsection{Spatial Activation Image View} The 3D image input is obtained by the voxel-wise time average of the fMRI time series, producing a mean activation map used as the intensity profile. Although it retains only time-averaged intensity, the 3D mean image is fast to compute and lightweight.

Although ROI-based functional brain graphs are successfully used in fMRI analysis~\cite{braingnn,zhao-stgnn,stnagnn}, they primarily encode relative inter-regional connectivity and ignore absolute signal intensity. In contrast, the time-averaged 3D fMRI volume preserves voxel-wise intensity and provides a complementary spatial view of the data. This complementary representation itself has been successfully applied to tasks such as machine-learning-aided age prediction and ASD classification~\cite{cnn-age,braincnn}. In applications such as chronological age prediction, voxel-wise features have also been shown to outperform coarse ROI-based inputs~\cite{age-whole-brain1, age-whole-brain2}. In our experiments, we find the inclusion of 3D image intensity information crucial for the performance of representation learning on fMRI.

\section{Method}
\subsection{Notation and Problem Definition}
Assume a task set $\mathcal{J}$ containing all distinct fMRI tasks. For each fMRI time series from subject $i$ and fMRI task $j\in \mathcal{J}$, we form an ROI-level functional brain graph $G_{i,j}$ and a 3D mean-intensity image $I_{i,j}$. A graph with $n$ ROIs has a vertex set $V_{i,j}$ of node features and an edge set $E_{i,j}$ with weighted edges. An input instance is
\[
x_{i,j}=(I_{i,j},\,G_{i,j}), \quad G_{i,j}=(V_{i,j},E_{i,j}).
\]
Our goal in pretraining is to learn an encoder $f\colon x\rightarrow\mathbb{R}^d$ that maps each data instance to a compact vector representation $z_{i,j}\in\mathbb{R}^d$,
\[
z_{i,j}=f(x_{i,j}).
\]
For downstream classification and regression, we attach a goal-specific head $g$ to the encoder $f$ either frozen or fine-tuned taking the learned representation $z$ as input. Let $y_{i,k}$ denote the goal-specific label of subject $i$ on a predictive task $k$, %In classification tasks, we optimize $g$ for maximum prediction probability in the label class. In regression, we minimize the difference between prediction output and the ground truth label.  
\[
y_{i,k}=g_k(z_{i,j})\quad \forall j\in \mathcal{J}.
\]

\subsection{BrainSimSiam Architecture}

Fig.~\ref{fig_encoder}a illustrates the general SimSiam~\cite{simsiam} framework. For each input instance $x$, we sample two augmented views $x_1$ and $x_2$. The shared encoder $f$ and predictor $h$ map the input to the learned representations $z_1$ and $z_2$ and their predictor outputs $p_1$ and $p_2$, respectively. Training minimizes a symmetric cosine-similarity loss on the encoder and predictor outputs (Eq.~\ref{eq:simsiam_loss}). 
\begin{equation}
\label{eq:simsiam_loss}
\mathcal{L}_{cos}(x_1,x_2) = -\frac{1}{2}\cdot\frac{\mathbf{p_1} \cdot \mathbf{z_2}}{\|\mathbf{p_1}\| \|\mathbf{z_2}\|}-\frac{1}{2}\cdot\frac{\mathbf{p_2} \cdot \mathbf{z_1}}{\|\mathbf{p_2}\| \|\mathbf{z_1}\|}
\end{equation}
During backpropagation, gradients are stopped at $z$. The predictor module and stop-gradient operation jointly mitigates model collapse during training when using only positive sample pairs.

As shown in Fig.~\ref{fig_encoder}b, the BrainSimSiam encoder comprises two branches: a GNN graph encoder and a convolutional neural network (CNN) image encoder. The GNN branch uses two GAT graph convolution layers~\cite{gat} with global average and global max pooling after each layer as permutation-invariant readouts. The CNN branch has 4 layers of 3D convolution with kernel size 3 and ReLU activation. Embeddings from the GNN and CNN branches are concatenated and passed to the projection MLP to produce the representation $z$. 

\begin{figure*}
\centering
\includegraphics[width=\textwidth]{ 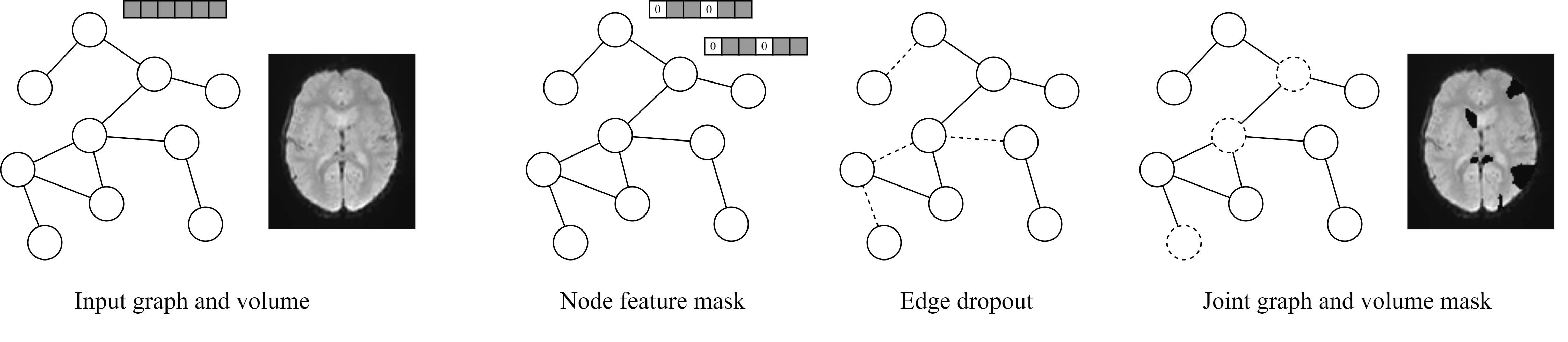}
\caption{Illustrations of augmentation operators used in training and interpretation.} \label{fig_augmentation}
\end{figure*}

\begin{figure*}
\centering
\includegraphics[width=\textwidth]{ 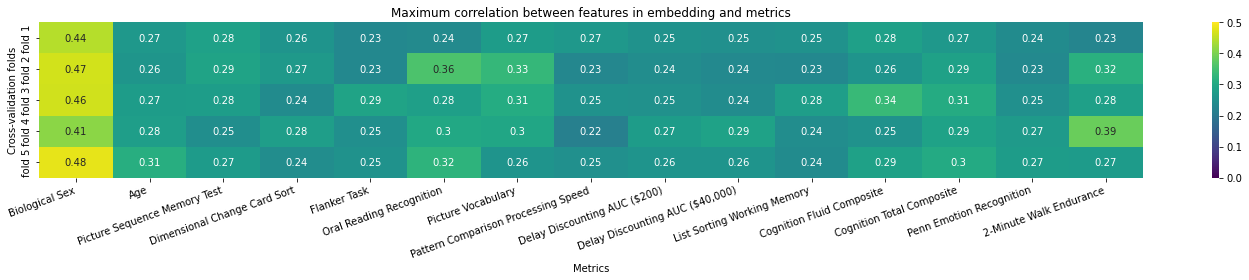}
\caption{Heatmaps of maximum absolute correlation of any single feature channels of learned embedding with metrics.} \label{fig_corr}
\end{figure*}

\subsubsection{ROI-aligned Graph-image Masking}
\label{sec:roi_mask}

In graph contrastive learning, it is common to compose random node feature masking and edge dropout to sample augmented views of graphs~\cite{cl-graph}. These operations, however, act only on the graph and leave the paired image data unchanged. To jointly perturb each ROI node and its corresponding region in the 3D image, we introduce an ROI-aligned graph-image masking operator that occludes the voxels in the image mapped to the dropped nodes in the ROI graph. Illustrations of each augmentation operator are shown in Fig.~\ref{fig_augmentation}. 

During training, we jointly apply node-feature masking, edge dropout, and the ROI-aligned graph-image masking operator. In quantitative results, we evaluate BrainSimSiam with or without ROI-aligned masking under identical training and evaluation schemes. Similarly, for post hoc interpretation with GNNExplainer\cite{gnnexplainer}, we use a soft variant of the ROI-aligned operator in conjunction with the soft node mask. 

Although evaluated in this paper only on paired graph and image views of fMRI, the ROI-aligned graph–image masking operator can generalize to multimodal settings, facilitating consistent alignment of graph-structured brain networks with registered complementary images from modalities like T1-weighted or diffusion MRI. 

\subsubsection{Task Invariance Loss}
\label{sec:task_loss}
Beyond the standard augmentation invariance loss, we introduce a task invariance loss for specific application to task-based fMRI. For each fMRI instance, we randomly sample a scan from the same subject under a different task and encourage the representation to preserve spontaneous, task-invariant functional patterns. Let $x_2^*$ denote an augmented view of the sampled scan from the same subject under a different task. The overall loss function  can be denoted as,
\begin{equation}
\label{eq:total_loss}
Loss = \mathcal{L}_{cos}(x_1, x_2) + \mathcal{L}_{cos}(x_1, x_2^*)
\end{equation}
where $\mathcal{L}_{cos}$ follows the definition of symmetric cosine similarity loss in Eq.~\ref{eq:simsiam_loss}.

The overall BrainSimSiam pipeline is shown in Fig.~\ref{fig_pipeline}, comprising the self-supervised pretraining described above and two downstream configurations: an MLP probe and end-to-end fine-tuning. The MLP probe offers a parameter-efficient, low-rank adaptation, whereas end-to-end fine-tuning updates the entire model at higher computational cost.

\subsection{Baselines}
We evaluate baselines for both classification and regression tasks on both the HCP and the Biopoint dataset under three settings: supervised learning, self-supervised learning (SSL) with MLP probing, and SSL with full fine-tuning.

\subsubsection{Supervised Learning}
To keep inputs comparable across methods, we pair widely used GNN backbones—GCN~\cite{gcn}, GAT~\cite{gat}, and GraphSAGE~\cite{sage}—with a parallel CNN encoder. For each downstream task, we train a separate model from scratch with cross-entropy loss or MSE loss for the classification or regression task, respectively. 

Although the supervised baselines forgo SSL pretraining, their training cost scales roughly linearly with the number of downstream tasks, since each task requires a separate model training run.

\subsubsection{Self-Supervised Learning with MLP Probing}
For a representative graph contrastive learning baseline, we adopt an existing design named Contrastive functional connectivity Graph Learning (CGL)~\cite{contrastive-population}, which forms positive and negative pairs and optimizes the SimCLR objective~\cite{simclr}. We report results for the original CGL implementation using graph input only~(CGL) and a variant implementation with a parallel CNN branch~(CGL-CNN) to allow for identical input with the proposed BrainSimSiam approach. In this setting, we pretrain the encoder once per dataset with the self-supervised loss, freeze the encoder weights, and train only a task-specific MLP head for each downstream classification or regression task. 

In this MLP probing setting, the cost of training an MLP head is negligible relative to the SSL pretraining. Thus, once the encoder is pretrained for each dataset, the total computation is approximately constant with respect to multiple downstream tasks. 

\subsubsection{Self-Supervised Learning with End-to-End Tuning}
We follow the same self-supervised pretraining models as in the MLP probing experiment. After pretraining, we initialize the model with the pretrained encoder, and then fine-tune all parameters in the encoder and MLP head jointly for each downstream task.The computational cost exceeds MLP probing but remains well below full supervised training, as fine-tuning requires only a few epochs per task.

\section{Experiments}
\subsection{Training Details}
We train our BrainSimSiam model and baselines with the SGD optimizer, batch size 128, and learning rate $1\times10^{-5}$ for 100 epochs. During training, we apply a weight decay factor of $0.001$ and a step learning rate scheduler with $\gamma=0.5$ that decays every 20 epochs.

To generate augmented views of each input graph and image, we apply node-feature masking ($p=0.5$), edge dropout ($p=0.5$), and ROI-aligned graph–image masking ($p=0.1$), sampled independently. The representation dimensionality is 2048 for HCP and 8192 for Biopoint for optimal performance in each dataset. All experiments are run on a single NVIDIA A100 GPU.

\subsection{Qualitative Correlation Analysis}
\begin{figure}
\centering
\includegraphics[width=0.9\columnwidth]{ 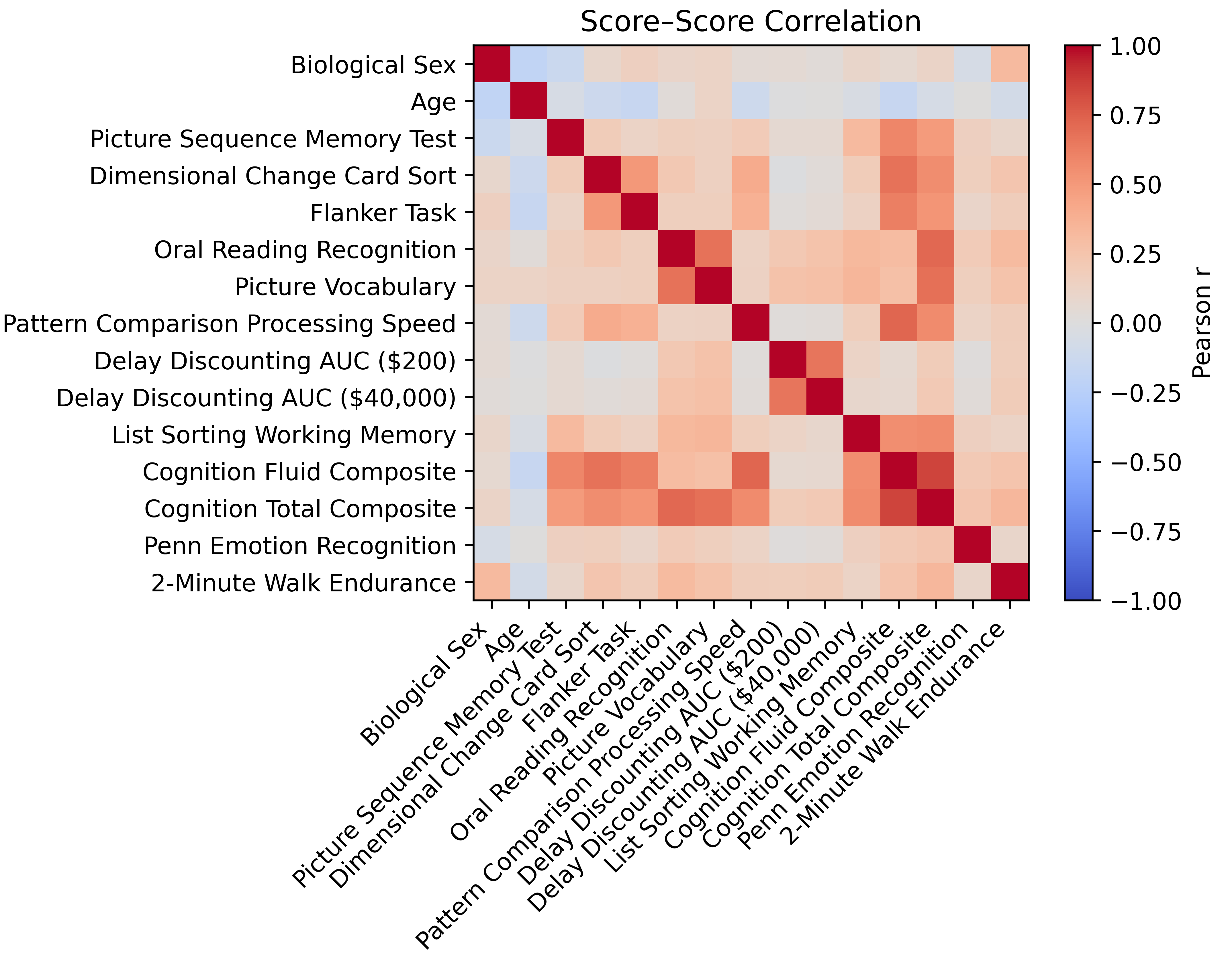}
\caption{Pearson's correlation between tested scores} \label{fig_score}
\end{figure}

In a zero-shot setting, we assess the quality of the BrainSimSiam learned embeddings on HCP for general-purpose prediction. Pearson's correlation between functional connectivity~(FC) and biological or behavioral phenotypes is widely applied in neuroscience for feature selection and subsequent statistical analysis~\cite{cpm}. The most correlated connections in the FC matrix are usually considered potential biomarkers or predictors for the corresponding target. We compare the association strength derived from our embeddings against FC elements to assess whether representation learning yields superior statistical signal to conventional FC.

As listed in Fig.~\ref{fig_score}, we select a set of 15 phenotypes including biological sex, age, physical endurance (2-minute walk endurance), and different measurements of cognitive function (row 3 to row 13) or emotion processing capability (Penn emotion recognition) from HCP. To justify the selection of tested phenotypes, we plot the score-to-score correlation between the 15 tested scores. The correlation heatmap shows high correlation only between a few cognitive measurements, indicating that the tested quantitative phenotypes are likely measuring heterogeneous aspects of human brain function related to fMRI signals. 

Using the encoder pretrained from self-supervision, we compute the embedding of data in the test set and correlate features of the embedding with the scores. For discrete classes, biological sex is labeled as 0 or 1. Age groups are assigned integer labels 0–3, ordered from youngest to oldest. In Table~\ref{tab_corr}, we show the maximum absolute correlation between any single feature channel of the learned embedding or any element in the FC matrix with all the 15 tested phenotypes. 

% table dim
\begin{table}[t]
\centering
\setlength{\tabcolsep}{1pt}
\begin{tabularx}{\columnwidth}{l  c   c}
\hline
\multirow{2}*{\textbf{Metric}}& Embedding & FC\\
{} & $r$ & $r$ \\
\hline
Biological Sex&                      \textbf{0.454} & 	0.273\\
Age& 	                             \textbf{0.278} & 	0.200\\
Picture Sequence Memory Test& 	     \textbf{0.274} & 	0.168\\
Dimensional Change Card Sort& 	     \textbf{0.258} & 	0.166\\
Flanker Task& 	                     \textbf{0.252} & 	0.176\\
Oral Reading Recognition& 	         \textbf{0.300} & 	0.197\\
Picture Vocabulary& 	             \textbf{0.294} & 	0.194\\
Pattern Comparison Processing Speed& \textbf{0.243} & 	0.159\\
Delay Discounting AUC (\$200)& 	     \textbf{0.256} & 	0.175\\
Delay Discounting AUC (\$40,000)& 	 \textbf{0.255} & 	0.173\\
List Sorting Working Memory& 	     \textbf{0.249} & 	0.184\\
Cognition Fluid Composite& 	         \textbf{0.283} & 	0.182\\
Cognition Total Composite& 	         \textbf{0.290} & 	0.211\\
Penn Emotion Recognition& 	         \textbf{0.254} & 	0.165\\
2-Minute Walk Endurance& 	         \textbf{0.297} & 	0.204\\
\hline
\end{tabularx}
\caption{Maximum absolute correlation of any feature channel from the learned BrainSimSiam embedding or any element in the functional connectivity matrix with the tested behavioral scores. Mean values from cross-validation are reported with higher value bolded.}
\label{tab_corr}
\end{table}

For each phenotype tested, the maximum absolute correlation of the feature in the learned embedding is consistently greater than that from FC. Additionally, in Fig.~\ref{fig_corr} we plot the maximum correlation of learned embedding with each metric for all five encoders trained in cross-validation experiments. It shows that the learned embedding has feature channels highly correlated with any of the tested metrics in all data partitions, with a minimum of $0.216$ across all cross-validation folds and metrics. On zero-shot, the learned embedding can generalize to the prediction of multiple phenotypes across different aspects of the brain function.

We show that the features from our learned fMRI representation are potentially better predictors for biological and behavioral phenotypes than functional connectivity. Combined with post hoc ROI-importance interpretation methods described in Section~\ref{sec:interpretation}, the learned fMRI features also retain similar interpretability to related ROIs compared to using FC.

% table HCP
\begin{table*}[t]
\centering
\setlength{\tabcolsep}{1.5pt}
\begin{tabularx}{\textwidth}{c  l  c c c  c c }
\hline
\multirow{2}{*}{\textbf{HCP}} & {} & \multicolumn{3}{c}{Biological Sex}& \multicolumn{2}{c}{Cog Score Composite}\\ 
{} & {} & Acc(\%) $\uparrow$ & F1 score $\uparrow$ & AUC $\uparrow$ & MAE  $\downarrow$ & $r$ $\uparrow$\\
\hline
\multirow{3}{*}{\textbf{Supervised}} & GAT-CNN & 85.7(1.7) & \textbf{0.846(0.030)} & 0.897(0.009) & \textbf{0.219(0.009)} & \textbf{0.153(0.072)} \\
& GCN-CNN & 83.5(1.9) & 0.826(0.026) & 0.887(0.019) & \textbf{0.219(0.007)} & 0.138(0.042) \\
& SAGE-CNN & \textbf{86.0(2.3)} & \textbf{0.846(0.034)} & \textbf{0.907(0.017)} & 0.222(0.009) & 0.135(0.060) \\
\hline
\multirow{4}{*}{\textbf{MLP Probe}} & CGL & 69.8(1.3) & 0.667(0.020) & 0.760(0.018) & 0.181(0.008) & 0.063(0.021) \\
& CGL-CNN & 85.4(1.1) & 0.841(0.013) & 0.928(0.013) & 0.178(0.004) & 0.070(0.036) \\
& BrainSimSiam & 86.6(1.2) & 0.854(0.023) &  0.933(0.004) & \cellcolor{table_2blue}0.170(0.004) & \cellcolor{table_1blue}\textbf{0.212(0.059)} \\
& BrainSimSiam w/ Mask & \textbf{87.0(1.7)} & \textbf{0.856(0.030)} & \textbf{0.937(0.012)} & \cellcolor{table_1blue}\textbf{0.169 (0.006)} & \cellcolor{table_2blue}0.207(0.061) \\
\hline
\multirow{4}{*}{\textbf{End-to-End}} & CGL & 73.0(2.8) & 0.727(0.029) & 0.826(0.017) & \textbf{0.172(0.007)} & 0.127(0.013)\\
& CGL-CNN & \cellcolor{table_2blue}90.5(3.1) & 0.885(0.053) & \cellcolor{table_1blue}\textbf{0.981(0.006)} & 0.176(0.004) & 0.187(0.040) \\
& BrainSimSiam & \cellcolor{table_2blue}90.5(2.2) & \cellcolor{table_2blue}0.892(0.035) & 0.979(0.012) & 0.174(0.006) & \cellcolor{table_1blue}\textbf{0.212(0.058)}\\
& BrainSimSiam w/ Mask & \cellcolor{table_1blue}\textbf{91.9(3.1)} & \cellcolor{table_1blue}\textbf{0.907(0.051)} & \cellcolor{table_2blue}0.980(0.010) & 0.178(0.007) & 0.198(0.070) \\
\hline
\hline
\multirow{2}*{\textbf{Biopoint}} & {} & \multicolumn{3}{c}{ASD}& \multicolumn{2}{c}{Age}\\ 
{} & {} & Acc(\%) $\uparrow$ & F1 score $\uparrow$ & AUC $\uparrow$ & MAE $\downarrow$ & $r$ $\uparrow$ \\
\hline
\multirow{3}{*}{\textbf{Supervised}} & GAT-CNN & \textbf{66.1(4.9)} & 0.699(0.049) &\textbf{0.694(0.096)} & 8.02(2.292) & 0.456(0.171) \\
& GCN-CNN & 64.6(5.7) & 0.640(0.121) & 0.589(0.050) & 8.09(0.573) & \textbf{0.542(0.201)} \\
& SAGE-CNN & 65.5(9.0) &\textbf{ 0.703(0.076)} & 0.620(0.133) & \textbf{7.66(1.361)} & 0.476(0.075) \\
\hline
\multirow{4}{*}{\textbf{SSL + MLP Probe}} & CGL & 59.8(3.1) & 0.689(0.041) & 0.571(0.053) & 2.97(0.358) & 0.364(0.041) \\
& CGL-CNN & 68.1(5.6) & 0.745(0.068) & 0.766(0.053) & 2.83(0.609) & 0.670(0.042) \\
& BrainSimSiam &\cellcolor{table_2blue}76.8(5.0) & \cellcolor{table_2blue}0.823(0.038) & \cellcolor{table_2blue}0.801(0.048) & 3.37(0.683) & 0.673(0.065) \\
& BrainSimSiam w/ Mask & \cellcolor{table_1blue}\textbf{79.1(4.9)} & \cellcolor{table_1blue}\textbf{0.843(0.039)} & \cellcolor{table_1blue}\textbf{0.813(0.072)} & \cellcolor{table_1blue}\textbf{2.11(0.591)} & \cellcolor{table_1blue}\textbf{0.773(0.073)} \\
\hline
\multirow{4}{*}{\textbf{SSL + End-to-End}} & CGL & 66.0(2.5) & 0.787(0.153) & 0.586(0.077) & 3.54(0.810) & 0.277(0.063) \\
& CGL-CNN & 69.5(2.7) & 0.791(0.158) & 0.652(0.114) & \cellcolor{table_2blue}\textbf{2.65(0.394)} & 0.624(0.127) \\
& BrainSimSiam & 68.8(4.3) & 0.794(0.026) & 0.673(0.096) & 2.84(0.147) & \cellcolor{table_2blue}\textbf{0.695(0.074)}\\
& BrainSimSiam w/ Mask & \textbf{72.1(4.1) }& \textbf{0.811(0.030)} & \textbf{0.688(0.095)} & 2.72(0.143) & 0.651(0.133) \\
\hline
\end{tabularx}
\caption{Performance of baseline models and BrainSimSiam on the HCP and Biopoint datasets. Metrics are reported as mean (std). For each dataset and experimental setting, the best result is bolded. Cell shading marks the top two results per dataset (darker for best, lighter for second best).}
\label{tab_result}
\end{table*}

\subsection{Quantitative Evaluation}
\begin{figure*}
\centering
\includegraphics[width=\textwidth]{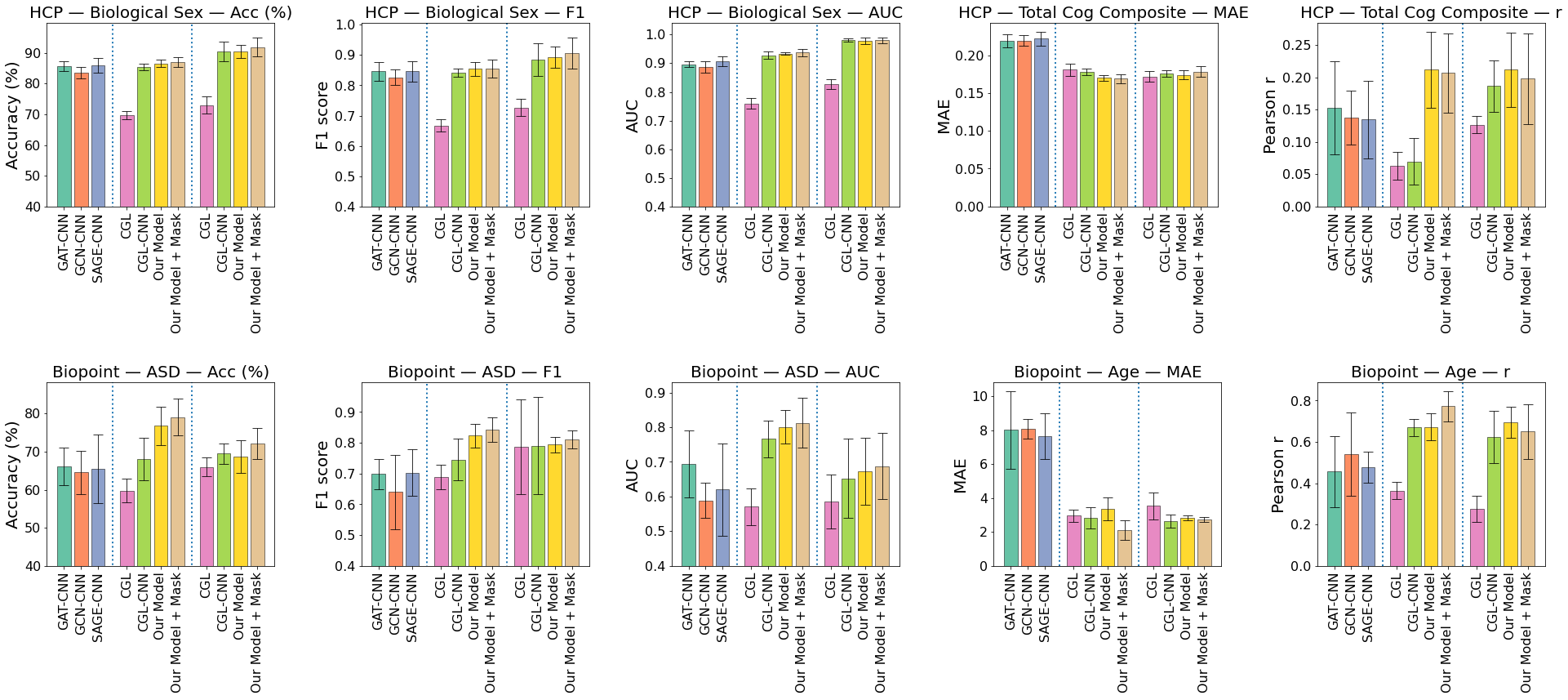}
\caption{Plots of quantitative performance on the classification and regression tasks from HCP and Biopoint dataset. Three training settings (Supervised, SSL + MLP Probe, SSL + End-to-End) are listed from left to right separated by blue dashed lines.} \label{fig_metric}
\end{figure*}

For quantitative evaluation, we select one classification and one regression task per dataset. On HCP, we evaluate biological sex classification and prediction of the total composite cognitive score. On Biopoint, we evaluate ASD classification and age prediction. For classification, we report accuracy (Acc), F1 score, and area under the ROC curve (AUC). For regression, we report mean absolute error (MAE) and Pearson correlation ($r$). For the BrainSimSiam framework, we show two implementations with or without applying the ROI-aligned graph-image masking during pretraining. Results are summarized in Table~\ref{tab_result}. To facilitate the interpretation of the table, Fig.~\ref{fig_metric} displays the metric performance of the baselines and our models in bar plots.

Across both datasets, BrainSimSiam model trained with ROI-aligned masking significantly ($p<0.05$) outperforms the supervised and self-supervised baselines. 
On HCP, where samples are abundant, supervised models approach the performance of self-supervised variants. Initializing from self-supervised pretraining and fine-tuning end-to-end yields the strongest performance on classification of biological sex at the cost of additional computation. In the regression task to predict total cognitive score composite, the performances in MLP-probing and end-to-end fine-tuning are similar. 
After fine-tuning, our model achieves 91.9\% accuracy and 0.980 AUC in biological sex classification task using only 52 million parameters (44 million in encoder and 8 million in MLP head), competitive with considerably larger fMRI foundation models like NeuroSTORM~(Acc 93.3\%, AUC 0.976)~\cite{neurostorm}. Compared to recent fMRI foundation models using only ROI-level input (BrainLM~\cite{brainlm} Acc 38.97\%, LCM~\cite{lcm} Acc 72.23\%), we observe a greater margin of improvement on the same classification task.

On Biopoint, with substantially fewer samples, the benefits of self-supervision are amplified. BrainSimSiam delivers more accurate and stable results for both classification and regression. Notably, MLP probing of the pretrained encoder surpasses end-to-end fine-tuning, indicating that freezing the encoder mitigates overfitting and highlighting the robustness of the learned representations. 

\subsection{Ablation Study}
For the ablation study, we evaluate the following model perturbations on both HCP and Biopoint using the baseline BrainSimSiam network and the training variation with ROI-aligned graph-image masking. All experiments are performed with SSL + MLP Probe setting. Mean values of metrics in cross-validation experiments are shown in Table~\ref{tab_ablation_base} and Table~\ref{tab_ablation_mask}. 
\subsubsection{W/O Projection MLP} Removing the projection MLP in encoder to mix information from GNN and CNN (Fig.~\ref{fig_encoder}b) substantially harms performance, suggesting that embeddings require nonlinearly integrating features from both the graph and the image view of fMRI rather than primarily relying on a single source of information.
\subsubsection{W/O Task Invariance Loss} We test removing the task-invariance loss described in Section~\ref{sec:task_loss}. The models have improved performance when trained on both the augmentation invariance loss and the task invariance loss.
\subsubsection{W/O GNN/CNN} We test removing the GNN or CNN module in the encoder (Fig.~\ref{fig_encoder}b), respectively. Similarly to removing the projection MLP in the encoder, removing the CNN module and corresponding input causes instability issues in training and leads to significant performance drops. 
\subsubsection{W/ Image Augmentation} For the variation of BrainSimSiam with ROI-aligned graph-image masking, we additionally compare the ablation on augmentation methods in the image space by training the model with additional 3D image augmentation, including random flipping, random rotation, and random intensity scaling and shifting. Results show that these standard augmentation methods used for tasks like image segmentation and denoising training harm the performance of the proposed self-supervised learning framework.

% table dim
\begin{table}[t]
\centering
\setlength{\tabcolsep}{2pt}
\begin{tabularx}{\columnwidth}{c | c c c | c c}
\hline
\textbf{\multirow{2}{*}{HCP}} & \multicolumn{3}{c|}{Biological Sex}& \multicolumn{2}{c}{Cog Score}\\ 
 & Acc(\%) & F1 score & AUC & MAE & $r$ \\
\hline
Model &\textbf{86.6} & \textbf{0.854} & \textbf{0.933} & \textbf{0.170} & \textbf{0.212}\\
\hline
W/o Proj MLP & 60.5 & 0.501 & 0.660 & 0.888 & 0.019\\
W/o Task Invar & 85.7 & 0.843 & 0.931 & 0.172 & 0.188\\
W/o GNN & 84.3 & 0.840 & 0.926 & 0.173 & 0.164\\
W/o CNN & 68.6 & 0.637 & 0.748 & 0.179 & 0.162\\

\hline
\hline
\textbf{\multirow{2}{*}{Biopoint}} & \multicolumn{3}{c|}{ASD}& \multicolumn{2}{c}{Age}\\ 
 & Acc(\%) & F1 score & AUC & MAE & $r$ \\
\hline
Model & \textbf{76.8} & \textbf{0.823} & \textbf{0.801} & 3.37 & \textbf{0.673}\\
\hline
W/o Proj MLP & 72.9 & 0.807 & 0.776 & 4.33 & 0.450\\
W/o Task Invar & 74.8 & 0.809 & 0.798 & 3.62 & 0.637\\
W/o GNN & 72.1 & 0.781 & \textbf{0.801} & \textbf{2.58} & 0.612\\
W/o CNN & 61.7 & 0.718 & 0.600 & 3.19 & 0.269\\
\hline

\end{tabularx}
\caption{Ablation study of BrainSimSiam on projection MLP, task invariance loss, and encoder backbone. Best performance in each column is bolded.}
\label{tab_ablation_base}
\end{table}

% table dim
\begin{table}[t]
\centering
\setlength{\tabcolsep}{2pt}
\begin{tabularx}{\columnwidth}{c | c c c | c c}
\hline
\textbf{\multirow{2}{*}{HCP}} & \multicolumn{3}{c|}{Biological Sex}& \multicolumn{2}{c}{Cog Score}\\ 
 & Acc(\%) & F1 score & AUC & MAE & $r$ \\
\hline
Model w/ Mask &\textbf{87.0} & \textbf{0.856} & 0.937 & \textbf{0.169} & \textbf{0.207}\\
\hline
W/o Proj MLP & 58.2 & 0.546 & 0.527 & 0.264 & -0.024\\
W/o Task Invar & 86.0 & 0.846 & \textbf{0.938} & 0.170 & 0.194\\
W/o GNN & 85.4 & 0.844 & 0.930 & 0.171 & 0.194 \\
W/o CNN & 68.7 & 0.666 & 0.748 & 0.182 & 0.152\\
W/ Image Aug & 77.4 & 0.752 & 0.856 & 0.170 & 0.143\\

\hline
\hline
\textbf{\multirow{2}{*}{Biopoint}} & \multicolumn{3}{c|}{ASD}& \multicolumn{2}{c}{Age}\\ 
 & Acc(\%) & F1 score & AUC & MAE & $r$ \\
\hline
Model w/ Mask & \textbf{79.1} & \textbf{0.843} & \textbf{0.813} & 2.11 & \textbf{0.773}\\
\hline
W/o Proj MLP & 73.8 & 0.812 & 0.772 & 5.56 & 0.281\\
W/o Task Invar& 76.0 & 0.820 & 0.799 & 2.26 & 0.763\\
W/o GNN & 70.9 & 0.776 & 0.787 & 2.37 & 0.764\\
W/o CNN & 64.4 & 0.745 & 0.597 & 2.82 & 0.410 \\
W/ Image Aug & 71.4 & 0.784 & 0.751 & \textbf{2.05} & 0.753 \\
\hline

\end{tabularx}
\caption{Ablation study of BrainSimSiam with ROI-aligned masking on projection MLP, task invariance loss, and encoder backbone. Best performance in each column is bolded.}
\label{tab_ablation_mask}
\end{table}

\begin{figure*}
\centering
\includegraphics[width=\textwidth]{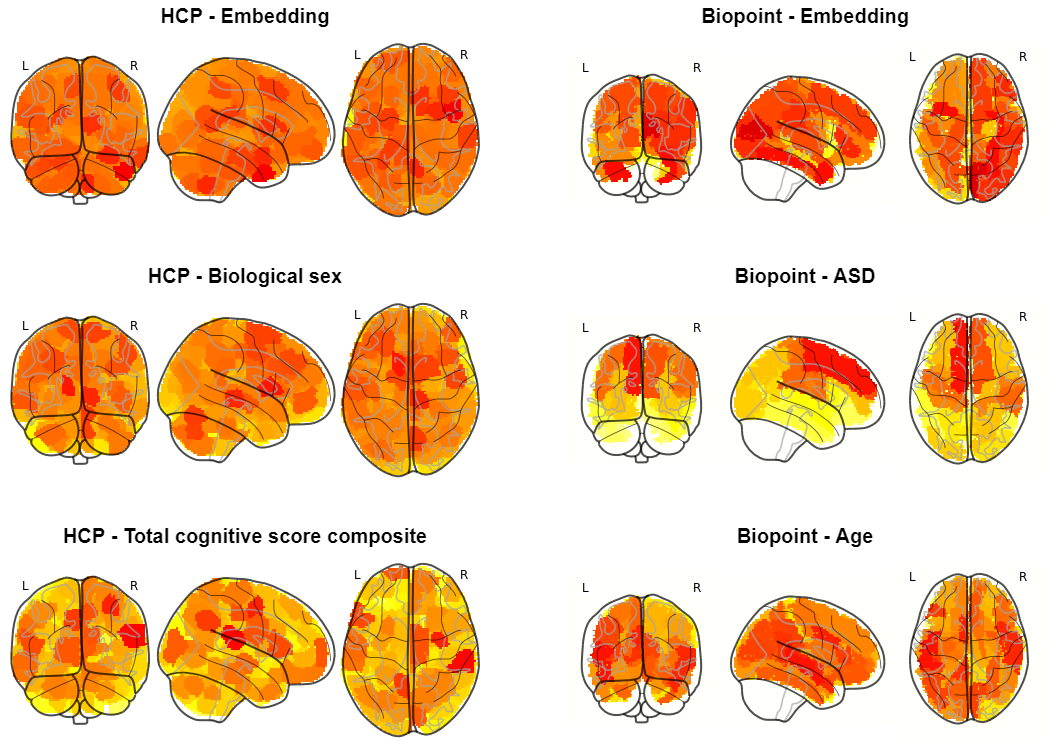}
\caption{Heatmaps of node importance in HCP and Biopoint. Darker regions indicate higher importance for generating embedding or prediction} \label{fig_interpret}
\end{figure*}

\section{Interpretation}
\label{sec:interpretation}
To explain the importance of brain ROIs for each specific task, we apply GNNExplainer~\cite{gnnexplainer} post hoc. On a graph level, GNNExplainer optimizes a learnable soft mask on node features and optimizes the mask to preserve the prediction of the model while promoting sparsity on the graph. In each iteration, it computes an element-wise product between node features and the mask and penalizes the difference between the original and masked prediction results by cross-entropy loss. 

In our joint graph–image input setting, we modify the implementation to mask paired graph and image input jointly. Since node features correspond to rows in FC matrices, each node feature channel corresponds to the connectivity information of one ROI to all other ROIs. Therefore, we extend GNNExplainer with a soft variation of the ROI-aligned mask described in Section~\ref{sec:roi_mask}. In optimizing a learnable soft mask for interpretation, we also map the node masks onto the corresponding voxels in the image, and those voxels corresponding to each ROI are soft-masked accordingly. 

For both datasets, we interpret ROI node importance on the encoder embedding and the classification/regression tasks respectively in the MLP probe setting. For pretrained models from 5-fold cross-validation experiments, we evaluate node importances on the testing set for each model and calculate an average node importance for each ROI and plot them in Figure~\ref{fig_interpret}. 

Across both datasets, the node importance scores for generating the general-purpose fMRI representation are relatively uniform across different ROIs, suggesting that pretraining learns embeddings from distributed whole-brain signals rather than from a small set of ROIs. In contrast, when applied on specific downstream tasks, the importance maps become more localized, highlighting salient regions important for generating task-specific predictions.

In the HCP dataset, the Shen atlas~\cite{shen268} partitions the brain via voxel-wise, spatially constrained spectral clustering and thus does not necessarily align with anatomical boundaries. Nevertheless, we find parcels proximal to the right thalamus to be relatively more informative for biological sex classification, and parcels near the right temporal gyrus to be strongly predictive of the total cognitive composite score. Both align with previous studies on sex differences~\cite{sex-dif} and non-verbal memory~\cite{non-verbal}.

In the Biopoint dataset, both the left and right superior frontal gyrus (SFG) exhibited higher importance for ASD classification. This observation aligns with previous research showing that atypical SFG connectivity is strongly associated with autism spectrum disorder~\cite{sfg-asd1,sfg-asd2}.
For the age prediction task, the superior temporal gyrus (STG) demonstrated greater importance, consistent with prior studies highlighting the STG’s critical role in the development of language and semantic processing~\cite{stg-age}. 

\section{Conclusion}
Recent advances in self-supervised learning have spurred foundation models across multiple application domains~\cite{att,dinov3}, including fMRI analysis~\cite{brainlm,neurostorm}. While large-scale pretraining in NLP and vision often improves transfer learning performance, evidence that pretraining on public fMRI datasets outperforms dataset-specific training remains limited—particularly when models trained on healthy cohorts are applied to disease-specific data without harmonization~\cite{fmri-harm}. 

We show that dataset-specific self-supervised pretraining yields robust, generalizable representations for fMRI, enabling accurate prediction of heterogeneous phenotypes. Across all downstream tasks, a BrainSimSiam encoder pretrained on the target dataset outperforms purely supervised baselines.

\bibliographystyle{elsarticle-num}
\bibliography{ref}
\end{document}